\documentclass{llncs}
\usepackage{graphicx}
\usepackage{latexsym}
\usepackage{amssymb}
\usepackage{latexsym}
\usepackage[pdftex]{hyperref}
\usepackage{adjustbox}
\usepackage[utf8]{inputenc}

\begin{document}

\title{Faster than LASER - Towards Stream Reasoning with Deep Neural Networks}

\author{
Jo\~ao Ferreira
\and 
Diogo Lavado
\and
Ricardo Gon\c calves
\and
Matthias Knorr
\and\\
Ludwig Krippahl
\and
Jo\~ao Leite
}

\institute{
NOVA LINCS \& Departamento de Inform\'atica, Universidade Nova de Lisboa
}

\maketitle

\begin{abstract}
With the constant increase of available data in various domains, such as the Internet of Things, Social Networks or Smart Cities, it has become fundamental that agents are able to process and reason with such data in real time. Whereas reasoning over time-annotated data with background knowledge may be challenging, due to the volume and velocity in which such data is being produced, such complex reasoning is necessary in scenarios where agents need to discover potential problems and this cannot be done with simple stream processing techniques. Stream Reasoners aim at bridging this gap between reasoning and stream processing and
LASER is such a stream reasoner designed to analyse and perform complex reasoning over streams of data. It is based on LARS, a rule-based logical language extending Answer Set Programming, and it has shown better runtime results than other state-of-the-art stream reasoning systems.
Nevertheless, for high levels of data throughput even LASER may be unable to compute answers in a timely fashion. In this paper, we study whether Convolutional and Recurrent Neural Networks, which have shown to be particularly well-suited for time series forecasting and classification, can be trained to approximate reasoning with LASER, so that agents can benefit from their high processing speed. 
\end{abstract}

\section{Introduction}
\label{intro}
We are witnessing a huge increase on data production in various domains, such as the Internet of Things (IoT), Industry 4.0, Social Networks and Smart Cities.
In order to deal with this highly dynamic data, to extract implicit relevant information from such data streams and be able to react to it in a timely fashion, agents need the ability to process, query and perform complex reasoning over such huge amounts of time-annotated data in real time.
For example, in an assisted living scenario, a robot designed to help elderly people who live alone should 
be able to use the data produced by several sensors to detect potential risk situations momentarily and react by calling for assistance in case of an emergency. Also, the data from different sensors in a Smart City scenario, such as traffic and pollution information, can help an agent to detect or even anticipate problems, and react to these by, for example, rerouting traffic to prevent traffic jams, specially on those areas with high levels of pollution, thus having an impact on the flow of traffic and the pollution footprint.

To deal with this highly dynamic data and their limited storage capacity for such data, agents can rely
 on stream processing systems~\cite{cugola2012processing}, i.e., Data Stream Management Systems (DSMS), which allow to continuously query on only a snapshot of a data stream,
and Complex Event Processors (CEP), which focus on the identification of temporal patterns occurring in a data stream. 
These systems cannot, however, perform complex reasoning tasks that require the integration of heterogeneous sources and background knowledge, 
such as, for example in the assisted living scenario, a database with the patient's records, a set of action policy rules, or a biomedical health ontology with information about diseases, their symptoms and treatments.

Stream Reasoning aims at overcoming these limitations, by bridging the gap between reasoning and stream processing~\cite{della2009s,dell2017stream}. One of the state of the art stream reasoners is LASER~\cite{BazoobandiBU17}, a system based on LARS~\cite{ref_lars}, which is a rule-based logical language extending Answer Set Programming (ASP)~\cite{GelfondL91}, designed to analyse and perform complex reasoning over streams of data. 
LASER allows one to declaratively encode a query as a set of logic programming rules, extending the language of ASP 
with temporal operators, such as $\lozenge$ (at some time instance) and $\square$ (at every time instance) and $@_t$ (at a specific time $t$), together with window operators, 
which allow focusing on portions of the data stream. 
Empirical results \cite{BazoobandiBU17} show that LASER has considerably better runtime results than other state-of-the-art stream reasoners, such as C-SPARQL~\cite{barbieri2010c} or CQELS~\cite{PhuocDPH11}.

Although LASER takes efficient computation into account while supporting expressive reasoning, such expressiveness necessarily imposes bounds on what can be computed efficiently. In fact, for high levels of data throughput and depending on the complexity of the considered queries, namely when the frequency of incoming data is higher than the time required for reasoning with that data, in particular when requiring to compute large combinatorics, LASER is not able to answer queries in time. 
This is prohibitive in real world scenarios where  the timely detection and reaction to problematic situations cannot be compromised.

Whereas Stream Reasoning systems, such as LASER, provide expressive sound and complete reasoning, 
it has been argued~\cite{HitzlerH10} that, for huge amounts of data, approximate methods need to be considered for obtaining answers in real time. 
Machine Learning approaches, such as Neural Networks~\cite{haykin1994neural}, have been successfully applied in a variety of domains for their ability to learn from examples, and generalize to unseen situations. Once trained, and since the computation of Neural Networks is only based on the manipulation of real-valued vectors and matrices, their processing speed is constant on the size of the input and usually extremely fast. 
Recent advances in Deep Learning~\cite{GoodfellowBC16} and its successful applications in several complex problems such as image, video, or speech recognition, led to the development of sophisticated Neural Networks capable of learning highly complex patterns from the data. This raised the interest of using such techniques to approximate formal deductive reasoning.

Within the field of neural-symbolic integration~\cite{BesoldEtAl17,HitzlerBES20} the problem of approximating logical reasoning with Neural Networks has been considered.
Early work on using neural networks for deductive reasoning opened the door for representing several propositional non-classical logics~\cite{GarcezLG2009}, building on systems that translate symbolic representations directly into neural networks~\cite{HammerH07,GarcezBRFHIKLMS15}.
Neural-symbolic approaches based on logic tensor networks~\cite{DonadelloSG17} learn sub-symbolic vector representations from training facts in a knowledge base. Other neural-symbolic approaches focusing on first-order inference include, e.g., CLIP++~\cite{FrancaZG14}, lifted relational neural networks~\cite{SourekAZSK18}, TensorLog~\cite{CohenYM20}, and recent work on approximating ontology reasoning~\cite{HitzlerBES20}.
Yet, none of these solutions considers reasoning with and reacting to  time-annotated, streaming data.

In this paper, we investigate the feasibility of using Deep Neural Networks to approximate stream reasoning with LASER, to take advantage of their high processing speed.
We explore two types of neural networks, namely Convolutional Neural Networks (CNN) \cite{TSC} and Recurrent Neural Networks (RNNs) \cite{qin2017dual}, which have been shown to obtain good results when applied to time-annotated data problems, such as time series forecasting and classification \cite{bianchi2018reservoir,fawaz2019deep}. For our experiments, we consider a real dataset with time-annotated sensor data on traffic, pollution and weather conditions, and explore different types of LASER queries, in order to cover different expressive features of its language. We obtain promising results, showing that it is indeed possible for agents to approximate reasoning with LASER with high levels of accuracy, and with significant improvements of the processing time. 
We also discuss the differences between the results obtained by the two types of neural networks considered.

In the remainder, we recall, in Section~\ref{prelim}, the necessary material on LASER and the two types of neural networks. 
Then, in Section~\ref{plan and methods}, we describe the methodology used for the experiments and, in Section~\ref{results}, the obtained results. Finally, in Section~\ref{discussion} we present a discussion of the results and draw some conclusions.

\section{Background}
\label{prelim}
In this section, we recall some useful notions for the remainder of the paper, namely those related to the stream reasoner LASER, and the two types of neural networks we consider, convolutional and recurrent neural networks.

\subsection{LASER}\label{ss:laser}
LASER~\cite{BazoobandiBU17} is a stream reasoning system based on LARS~\cite{ref_lars}, a rule-based logical language extending Answer Set Programming (ASP)~\cite{GelfondL91} with stream reasoning capabilities.  
The concept of data stream is modelled in LARS as a sequence of time-annotated atoms, more precisely, as pairs $S = \langle T,v \rangle$, where $T$ is a closed time interval timeline of natural numbers, and $v:\mathbb{N}\to 2^\mathcal{A}$ is an evaluation function, that indicates which atoms are true at each time point in $T$. 
The language of LARS contains the usual boolean operators, but offers also temporal operators such as $\lozenge$ and $\square$ to express that a formula holds, respectively, at some time or at every time of a given stream. The operator $@_t$ allows us to express that a formula holds at a specific time $t$ within a stream. An essential component of stream reasoning is the concept of \emph{window}, since it allows restricting the focus only to a portion of the stream, usually to the most recent elements. In LARS, tuple-based and time-based windows are considered.
A time-based window of size $n$, represented as $\boxplus^n$, restricts the focus only to those atoms that are true in the last $n$ time points of the stream, whereas a tuple-based window of size $n$, represented as $\boxplus^{\#n}$, restricts the focus to the last $n$ atoms of the  stream.

To address the trade-off between expressiveness and efficiency, LASER considers a tractable fragment of LARS, called Plain LARS. In more detail, given a set $\mathcal{A}$ of \emph{atoms}, the set $\mathcal{A}^+$ of \emph{extended atoms} is defined by the grammar
\[ a\ |\ @_t a\ |\ \boxplus^w@_t a\ |\ \boxplus^w\Diamond a\ |\ \boxplus^w\square a \]
where $a\in \mathcal{A}$, $t\in \mathbb{N}$ is a time point, and $w$ can be either $n$ or $\#n$.
Then, a Plain LARS program over $\mathcal{A}$ is a set of rules of the form
$\alpha \leftarrow \beta_1, \ldots, \beta_m$
where $\alpha$ is either $a$ or $@_t a$, with $a\in\mathcal{A}$, and each $\beta_i$ is an extended atom.

The resulting language is highly expressive allowing sofisticated queries for reasoning with streaming data as well as the direct encoding of rich background knowledge to be taken into account in the reasoning process. 

The semantics of Plain LARS programs is rooted in the notion of \emph{answer stream}, an extension of the answer set semantics for logic programs \cite{BazoobandiBU17}.

A fundamental feature of LASER is the implementation of an incremental evaluation procedure, which, by annotating formulae and efficiently propagating and removing such annotations, avoids many unnecessary re-computations.

\subsection{Neural Networks}
We now present a brief overview of the two types of neural networks considered here, 
for which we assume familiarity with the basic notions on supervised learning with Artificial Neural Networks~\cite{GoodfellowBC16,haykin1994neural}.

\subsubsection{Recurrent Neural Networks} 
Recurrent Neural Networks (RNNs) form a class of Artificial Neural Networks that have shown the ability to detect patterns in sequential data \cite{qin2017dual}. RNNs have been successfully applied to speech recognition and synthesis and time-series prediction and classification. The key distinguishing feature of this type of architecture is how the information passes through the network. They have a feedback mechanism, in which the output of the recurrent layer gets transmitted back into itself in the next step. This allows the network to take into account previous inputs, besides the current one. 
Long Short Term Memory (LSTM)~\cite{hochreiter1997long} networks are a type of RNN built from gated cells, structured blocks that allow for a better control of the information flow over the network.
 These gated cells receive as input the data from the current time step, the hidden state and cell state from the previous time step. The new hidden state will depend on these three inputs after nonlinear transformations with activation functions. The new cell state, however, will only be subject to linear transformations. This helps preventing the vanishing gradients problem~\cite{Hochreiter98} and preserving long-term information through the network, improving the identification of patterns extendning over longer periods of time.

\subsubsection{Convolutional Neural Networks}
Convolutional Neural Networks (CNNs) are well-known for their successful application in processing data with positionally invariant patterns, such as images, where CNNs are the state of the art methods, or time series, which can be seen as one-dimensional grids of data organized along the time axis~\cite{ref_cnn02}. CNNs include convolutional layers in which kernels are applied to the input data in order to produce feature maps capable of detecting useful patterns. These patterns are then fed into subsequent layers to help identify increasingly complex features~\cite{ref_cnn04,ref_cnn02}.   
In addition, CNNs include pooling layers that aggregate patches of feature maps to help abstract from the position of detected features in the input and reduce the dimension of the data. The use of these two kinds of layers leads to a significant decrease in the number of trainable parameters compared to using only fully connected layers. Nevertheless, CNNs used for classification problems generally include a small set of fully connected layers at the end, after convolution and pooling layers extracted useful features form the data. Dropout layers can also be used for regularization. Dropout layers randomly set to zero outputs from the previous layer at each batch during training. This helps reduce overfitting by forcing the network to avoid relying too much on specific inputs instead of broader patterns.

\section{Methods}
\label{plan and methods}
In this section, we present the experiments we designed to test whether stream reasoning with LASER can be approximated by RNNs and CNNs.

In general terms, each experiment consisted of $i)$ the development of a LASER query, considering different aspects of the expressiveness of its language as detailed in Section~\ref{ss:laser}, and its execution in order to obtain the results according to LASER; $ii)$ the use of such labeled data (an encoding of the input data stream and the corresponding output according to LASER) to train and test different RNN and CNN models, to be able to choose, for each of these types, the best model in terms of approximating stream reasoning with LASER.

Before we detail these two steps, we introduce the considered dataset.

\subsection{Dataset}
The dataset we used in our experiments is a publicly available dataset\footnote{\url{http://iot.ee.surrey.ac.uk:8080/datasets.html}}
 containing time-stamped information about road traffic, pollution, and weather reports obtained from 449 sensors distributed  throughout the city of Aarhus in Denmark. The dataset includes, for example, \textit{pollution} events, such as ozone, carbon monoxide, sulfure dioxide and nitrogen dioxide measures, on average 1942 pollution events per timestamp of five minutes, \textit{traffic} events such as vehicle count and average speed, on average 700 traffic events per timestamp, and \textit{weather} events such as wind speed and direction, temperature, humidity and pressure measures, on average 2 weather events per timestamp. 

To allow for more complex queries involving advanced reasoning, we grouped the sensors into sectors, according to their geographic position, thus enabling comparisons between sectors and the sensors within these. We also considered rich background knowledge, namely the definition of static notions such as city, town and suburbs of cities, and relations between these, such as adjacency.

Taking into account the mentioned limitation of LASER to process large amounts of data, 
we reduced the number of sensors to be considered from 499 to 273, which also allowed for a more even distribution of sensors per sector.
The data from the dataset provided in .csv was encoded as facts indicating the occurrence of an event in a sector, associated to the time interval as natural numbers, where each time step corresponds to five minutes.
The resulting data set contains a total of 17569 time steps that was used as input for Laser.

\subsection{LASER queries}
We developed a set of queries to systematically explore different aspects of LASER's expressive language, namely the usage of combinations of window operators $\boxplus$ and temporal operators $\square$, $\Diamond$, and $@_t$.
For part of these queries, we enriched the reasoning process with background knowledge, i.e., knowledge that is not derived from the stream, including notions such as adjacency, an ontology over the concepts of city, town and suburb, and topological information, such as which towns are suburbs of which cities.
 Given that neural networks can only provide answers through a fixed number of output neurons, we only considered LASER queries with a fixed number of answers, i.e., boolean or discrete. For boolean queries, we took into account the overall balance between positive and negative outputs, since highly unbalanced scenarios could impose unwanted bias in the training process of the neural networks. 
For queries with discrete output, the final count was obtained by a simple post-processing step, as LASER does not include aggregation functions.
To construct and process LASER queries, we built on the available prototype of LASER \footnote{\url{https://github.com/karmaresearch/Laser}} extended with means to represent factual background knowledge more straightforwardly and with additional operators to facilitate the comparison of strings and numeric values.\footnote{\url{https://github.com/jmd-ferreira/approx_laser}.}

\subsection{Training and testing CNNs and RNNs}
For each query, the neural networks considered receive an encoding of the input stream, just as the query, and produce an output of the same type as the output of the query. Such outputs are sequences of time steps, similar to streams, containing the query results, and are used as labels to train the neural networks. To encode the input data for the neural network, we use an array of fixed size corresponding to the sensor readings that are relevant to the query, structured w.r.t.\ sectors and events.
In the case of LSTMs, the RNNs considered here, an input is a sequence of size $w$, where $w$ is the window size  used for the query, and each element of that sequence is an array containing the features per sector. 
For CNNs, the input is a matrix of size $n\times w$, corresponding to the entire sequence of $w$ arrays, each of size $n$.
We also used standardization, to account for range differences between features, to prevent a slow or unstable learning  processes.

For each query, we designed, trained and tested different architectures, both for CNNs and LSTMs, varying, for example, the number and type of layers, the number of neurons in each layer, or the kernel size for CNNs. We also used Dropout layers after dense layers to avoid overfitting. The training and test set accuracy (the fraction of correct answers)  of these neural networks were used as a measure to assess their ability to approximate LASER. 
In real applications, one must often consider that different errors have  different consequences and so other metrics may be useful, such as confusion matrices, AUC, precision and recall or the F-score. However,  since our goal is to assess how well the networks approximate LASER in a range of hypothetical scenarios chosen to span different types of queries and problems, we decided to focus on accuracy. 

The available 17569 samples were divided into 80\% used for training (with either 30\% or 20\% reserved for training validation depending on the query) and 20\% for testing.
The number of epochs for training varies for the queries, namely, for LSTMs, between 100 and 200 and, for CNNs, between 50 and 200. 

The design, training and testing of our CNNs and RNNs models was done using the Python library Tensorflow\footnote{https://blog.tensorflow.org/2019/09/tensorflow-20-is-now-available.html} and Keras\footnote{{https://www.tensorflow.org/guide/keras}}, a high-level API for Tensorflow that provides implementations of CNNs and LSTM architectures. 
\section{Description of Experiments and Results}
\label{results}
In this section, we present the experiments done to assess whether LASER can be approximated by RNNs and CNNs. We considered a set of queries exploring different aspects of LASER's expressive language (cf. Section~\ref{ss:laser}), along with different types of problems, such as classification or regression, and the use of background knowledge, and we present representatives for such different queries.
For each experiment, we present the considered query and discuss the results in terms of approximation obtained by the CNNs and LSTMs, the considered type of RNNs. To ease readability, we opted to present just a natural language description of each query, instead of its formal representation in LASER. 
 \footnote{The full specification of the queries, the final configurations of the networks and graphics of the learning phase can be found in the Appendix.} 



The experiments were run on a computer with an Intel i7 8th gen hexa-core processor, with GPU Nvidia 1050 Ti, and 
8GB of RAM.
We note that pre-processing the input data (for LASER) as well as encoding the input for the networks took only a few milliseconds per time step, which is why it is not reported individually, and that training the networks required on average not more than ten minutes for both kinds of networks, which is only once before applying the network, and, once trained, neural networks never required more than 300 $\mu s$ to produce the output for one time step in any of the experiments.

\subsection{Test Case 1.}
The first experiment aimed at considering a complex temporal boolean query, leveraging on LASER's expressive temporal language.
This query considers intermediate notions of industrial, urban and highway events, defined using different temporal patterns of pollution and traffic events. 
A metropolitan event occurs in a sector within a given time window if at least one industrial event, one urban event, and one highway event are detected in that sector in that time window.
The results for the following query are reported in the subsequent table: 

``\textit{Are there any metropolitan events in the last 9 time steps?}"

\begin{center}
\begin{adjustbox}{max width=0.8\textwidth}
\begin{tabular}{ c | c | c | c | c| }
\cline{2-5}
 & LSTM Train Acc & LSTM Test Acc & CNN Train Acc & CNN Test Acc  \\
\hline
\multicolumn{1}{|c|}{Query 1}& 0.9667& 0.9107 & 0.9951 & 0.9996 \\
\hline
\end{tabular}
\end{adjustbox}
\end{center}

During the training phase, we observed that LSTM models showed considerable levels of overfitting and some inconsistency in terms of accuracy and loss, which did not occur with CNNs, whose performance was more stable.
Still, regarding the test set, both types of networks showed excellent results, with those of CNNs being slightly better than those of RNNs.

Overall, the results obtained allow us to conclude that it is indeed feasible to consider both CNNs and RNNs to approximate the complex temporal reasoning of LASER on boolean queries. Moreover, the gain in terms of processing time is impressive, as the reported limit of 300$\mu s$ to produce the output for each time step, is several orders of magnitude faster than LASER's average of 89 seconds per time step. Also note that LASER's processing time here becomes prohibitive when the data in this scenario arrives with a frequency of less than a minute.

\subsection{Test Case 2.}

In the second experiment, we considered a classification problem that involves temporal reasoning and complex comparisons between sectors. The aim was to test whether neural networks are able to approximate stream reasoning when the problem is not just a binary classification problem, as in the previous test case, but rather a complex multi-label classification problem.  

This query involves temporal concepts such as high and low traffic and pollution, which depend on different traffic and pollution measures given by the sensors. 
Two sectors are then called conflicting with respect to traffic and pollution if they consistently differ on the registered traffic and pollution events within a window of three time steps, 
meaning that one of the sectors always registered high traffic and high pollution, and the other always low traffic and low pollution.
The query and the obtained results are given as follows:

``\emph{Are there any pairs of conflicting sectors within the last 3 time steps?}"

\begin{center}
\begin{adjustbox}{max width=0.8\textwidth}
\begin{tabular}{ c | c | c | c | c| }
\cline{2-5}
 &  LSTM Train Acc & LSTM Test Acc & CNN Train Acc & CNN Test Acc  \\
\hline
\multicolumn{1}{|c|}{Query 2}& 0.9997& 0.9896 & 0.9956 & 0.9951 \\
\hline
\end{tabular}
\end{adjustbox}
\end{center}

During the training phase, and contrarily to CNNs, RNNs showed slight signs of overfitting on the last epochs. Nevertheless, we can see that both CNNs and RNNs present excellent results for this query. 
Hence, multi-label classification problems in stream reasoning can also be approximated with CNNs and LSTMs.
Here, LASER requires on average 10 seconds per time step, which is still slower than processing with neural networks by a large margin.

\subsection{Test Case 3.}

In this experiment, we considered a problem involving temporal reasoning and complex classification of sectors. The aim was to test if neural networks can approximate stream reasoning for complex multiclass classification problems.  

This query considers complex temporal conditions on pollution and traffic, to define urban, work, rural and industrial events. Then, for a given time window, a sector is classified as rural if at least one rural event was registered during that time window; as a living sector if only urban events were registered during that time window; as an office sector if both urban and working events were registered during that time window, and no other type of event was registered; and as a factory sector if industrial events were registered in that time window, but no rural events. The query and the results are as follows:

``\textit{Based on the last 3 timestamps, what is the classification of each sector?}"

\begin{center}
\begin{adjustbox}{max width=0.8\textwidth}
\begin{tabular}{ c | c | c | c | c| }
\cline{2-5}
 &  LSTM Train Acc & LSTM Test Acc & CNN Train Acc & CNN Test Acc  \\
\hline
\multicolumn{1}{|c|}{Query 3}& 0.9996& 0.9962 & 0.9974 & 0.9933 \\
\hline
\end{tabular}
\end{adjustbox}
\end{center}

This query, which defines a multiclass classification problem, makes extensive use of LASER's expressive power. Since the classification of each sector is independent of the others, we trained ten neural networks, one per sector, each providing the classification for one sector, 
which resulted in much better approximation results than considering just one neuronal network for all sectors.

We can see that both CNNs and RNNs present excellent results for this query, thus allowing us to conclude that CNNs and RNNs can also approximate stream reasoning for the case of multiclass classification problems. 
Processing with LASER took on average 30 seconds per time step which is again much slower than processing with Neural Networks.

\subsection{Test Case 4.}

In this experiment, we considered a problem that involves temporal reasoning, complex classification and counting of sectors. The aim was to test whether neural networks are able to approximate stream reasoning when the problem is a complex regression problem.
This query considers the notion of rural event, which is detected in a sector if, in a given time window, that sector registers low and decreasing measurements of traffic and pollution. 
The considered query is:

``\textit{In the last four time steps, how many sectors registered a rural event?}" 

For this regression problem, we considered the mean squared error as a measure of approximation, and 
the results are reported in the following table.

\begin{center}
\begin{adjustbox}{max width=0.8\textwidth}
\begin{tabular}{ c | c | c | c | c| }
\cline{2-5}
 &  LSTM Train Loss & LSTM Test Loss & CNN Train Loss & CNN Test Loss  \\
\hline
\multicolumn{1}{|c|}{Query 4}& 0.05& 0.13 & 0.01 & 0.006 \\
\hline
\end{tabular}
\end{adjustbox}
\end{center}

We can observe very goods results for LSTMs, and excellent ones for CNNs. 
This is as clear indication that both CNNs and RNNs can approximate stream reasoning for the case of regression problems. 
In this experiment, LASER required on average 7 seconds per time step for answering the query, which is again a lot slower than network processing. 

\subsection{Test Case 5.}

In this experiment, we considered a problem that involves the combination of temporal reasoning with background knowledge.
The aim was to test whether neural networks are able to approximate stream reasoning when temporal reasoning over the stream is enhanced with rich background knowledge.

For this query we considered LASER's ability to define and reason with background knowledge, and defined concepts such as city and town, with towns being suburbs of a city, and a proximity relation between suburbs based on whether these are suburbs of the same city.
In addition, we defined a notion of anomaly occurrence in a sector in a given time window, if such sector is a city (static classification) and it is classified as an industrial sector (based on the occurrence of temporal events), together with two of its suburbs.
To cope with the extra complexity of this query, and to obtain a balanced training set, we only considered the measures given by a subset of 100 sensors. 
The query and the obtained results are the following:

``\textit{In the last four timestamps, in which sectors an anomaly has been detected?}" 

\begin{center}
	\begin{adjustbox}{max width=0.8\textwidth}
		\begin{tabular}{ c | c | c | c | c| }
			\cline{2-5}
			& LSTM Train Acc &LSTM Test Acc & CNN Train Acc & CNN Test Acc  \\
			\hline
			\multicolumn{1}{|c|}{Query 5}& 0.9628& 0.9418 & 0.9519 & 0.9341 \\
			\hline
		\end{tabular}
	\end{adjustbox}
\end{center}

We can see that
both types of networks present excellent results, again showing clear benefits in terms of processing time for the networks.
This experiment further strengths the hypothesis that both CNNs and RNNs can be successfully used to approximate stream reasoning with LASER, even when complex temporal reasoning over the stream is combined with rich background knowledge.

\section{Conclusions}
\label{discussion}
In this paper, we have investigated the viability of using Recurrent and Convolutional Neural Networks to approximate expressive stream reasoning with LASER, so that agents can leverage on their fast processing time, which would make them suitable for scenarios where, due to the velocity and size of data to be processed, LASER cannot be used to provide answers in a timely fashion.   

Our experiments on a real dataset with time-annotated sensor data on traffic, pollution and weather conditions, show that both RNNs and CNNs provide excellent approximations to stream reasoning with LASER, even for different types of complex queries exploring LASER's expressive language,  providing at the same time incomparably better performance in terms of processing time.

In terms of comparing CNNs and RNNs results, we could not observe a significant difference between the two in terms of approximation. During the training phase, however, we observed that, contrarily to the case of CNNs, RNNs were usually more prone to overfitting and their accuracy would sometimes significantly oscillate between epochs. In general terms, however, we could not conclude that one of the architectures is better suited than the other for approximating stream reasoning, despite the small advantages of CNNs on queries 1 and 4.

We should note that LASER, which has shown promising results in terms of processing time and expressiveness when compared with other state-of-art stream reasoners, is still useful in scenarios where agents need certain answers, or when the low size and velocity of the incoming data allows LASER to present answers in time, or where not enough training data is available or is imbalanced.

For future work, it would be interesting to experiment with larger datasets, to see if it is possible to improve on those cases where the results were not excellent. Lastly, we have focused on the approximation of stream reasoning with neural networks to leverage on their high processing speed, but it would also be interesting to investigate, in a stream reasoning scenario, how agents can benefit from neural networks' natural ability to cope with noisy data.

\paragraph{Acknowledgments}
We thank the anonymous reviewers for their helpful comments and acknowledge support by FCT project RIVER ({PTDC}/{CCI-COM}/{30952}/{2017}) and by FCT project NOVA LINCS ({UIDB}/{04516}/{2020}). 

\bibliographystyle{splncs04}
\bibliography{biblio}

\newpage
\section{Appendix}
\label{appendix}
\subsection{LASER queries} 

\subsubsection{LASER code for Query 1}
\begin{verbatim}
poll_inc(MES,SEC) :- time_win(9, 0, 1, @(T, pollution(MES,VAL,SEC)))
  and time_win(8, 0, 1, @(T1, pollution(MES,VAL2,SEC)))
  and MATH(+,RT,T,1)  and COMP(==, RT, T1)
  and COMP(>=, VAL, VAL2) ,

traff_inc(MES,SEC) :- time_win(9, 0, 1, @(T, traffic(MES,VAL,SEC)))
  and time_win(8, 0, 1, @(T1, traffic(MES,VAL2,SEC)))
  and MATH(+,RT,T,1)  and COMP(==, RT, T1)
  and COMP(>=, VAL, VAL2) ,

poll_dec(MES,SEC) :- time_win(9, 0, 1, @(T, pollution(MES,VAL,SEC)))
  and time_win(8, 0, 1, @(T1, pollution(MES,VAL2,SEC)))
  and MATH(+,RT,T,1)  and COMP(==, RT, T1)
  and COMP(<=, VAL, VAL2) ,

traff_dec(MES,SEC) :- time_win(9, 0, 1, @(T, traffic(MES,VAL,SEC)))
  and time_win(8, 0, 1, @(T1, traffic(MES,VAL2,SEC)))
  and MATH(+,RT,T,1)  and COMP(==, RT, T1)
  and COMP(<=, VAL, VAL2) ,

traff_low(MES,SEC) :- time_win(9, 0, 1, diamond(traffic(MES,VAL,SEC)))
  and COMP(>=,VAL, 10)  and COMP(<=,VAL,11),

traff_high(MES,SEC) :- time_win(9, 0, 1, diamond(traffic(MES,VAL,SEC)))
  and COMP(>=,VAL, 250)   and COMP(<=,VAL,300),

poll_low(MES,SEC) :- time_win(9, 0, 1, diamond(pollution(MES,VAL,SEC)))
  and COMP(>=,VAL, 0)  and COMP(<=,VAL,15),

poll_high(MES,SEC) :- time_win(9, 0, 1, diamond(pollution(MES,VAL,SEC)))
 and COMP(>=,VAL, 214)  and COMP(<=,VAL,215),

industrial_area(SEC) :- time_win(9, 0, 1, diamond(poll_inc(MES, SEC)))
 and time_win(9, 0, 1, diamond(traff_dec(MES10,SEC)))
 and time_win(9, 0, 1, diamond(poll_high(MES1,SEC)))
 and time_win(9, 0, 1, diamond(traff_low(MES2,SEC)))
 and COMP(s!=, MES, MES10),

industrial_box(SEC) :- time_win(3, 0, 1, box(industrial_area(SEC))),

highway_area(SEC) :- time_win(9, 0, 1, diamond(traff_inc(MES, SEC)))
 and time_win(9, 0, 1, diamond(poll_inc(MES10,SEC)))
 and time_win(9, 0, 1, diamond(poll_high(MES2,SEC)))
 and time_win(9, 0, 1, diamond(traff_high(MES4,SEC)))
 and COMP(s!=, MES, MES10),

highway_box(SEC) :- time_win(3, 0, 1, box(highway_area(SEC))),

urban_area(SEC) :- time_win(9, 0, 1, diamond(traff_inc(MES, SEC)))
 and time_win(9, 0, 1, diamond(poll_dec(MES10,SEC)))
 and time_win(9, 0, 1, diamond(poll_low(MES2,SEC)))
 and time_win(9, 0, 1, diamond(traff_low(MES4,SEC)))
 and COMP(s!=, MES, MES10),

urban_box(SEC) :- time_win(3, 0, 1, box(urban_area(SEC))),

city(SEC) :- industrial_box(SEC)
 and highway_box(SEC)
 and urban_box(SEC),
\end{verbatim}

\subsubsection{LASER code for Query 2}

\begin{verbatim}
traff_low(MES,SEC) :- time_win(3, 0, 1, box(traffic(MES,VAL,SEC)))
 and COMP(<=, VAL, 45),

traff_high(MES,SEC) :- time_win(3, 0, 1, box(traffic(MES,VAL,SEC)))
 and COMP(>=, VAL, 150),

poll_low(MES,SEC) :- time_win(3, 0, 1, box(pollution(MES,VAL,SEC)))
 and COMP(<=, VAL, 16),

poll_high(MES,SEC) :- time_win(3, 0, 1, box(pollution(MES,VAL,SEC)))
 and COMP(>=, VAL, 195),

x(SEC1,SEC2) :- traff_low(MES1,SEC1)
 and poll_low(MES2,SEC1)
 and traff_high(MES1,SEC2)
 and poll_high(MES2,SEC2)
 and time_win(3, 0, 1, diamond(parking(MES,VAL,SEC1)))
 and COMP(!=, SEC1, SEC2)
\end{verbatim}

\subsubsection{LASER code for Query 3}

\begin{verbatim}
poll_inc(MES,SEC) :- time_win(9, 0, 1, @(T, pollution(MES,VAL,SEC)))
 and time_win(8, 0, 1, @(T1, pollution(MES,VAL2,SEC)))
 and MATH(+,RT,T,1)  and COMP(==, RT, T1)
 and COMP(>=, VAL, VAL2) ,

traff_inc(MES,SEC) :- time_win(9, 0, 1, @(T, traffic(MES,VAL,SEC)))
 and time_win(8, 0, 1, @(T1, traffic(MES,VAL2,SEC)))
 and MATH(+,RT,T,1)  and COMP(==, RT, T1)
 and COMP(>=, VAL, VAL2) ,

poll_dec(MES,SEC) :- time_win(9, 0, 1, @(T, pollution(MES,VAL,SEC)))
 and time_win(8, 0, 1, @(T1, pollution(MES,VAL2,SEC)))
 and MATH(+,RT,T,1)  and COMP(==, RT, T1)
 and COMP(<=, VAL, VAL2) ,

traff_dec(MES,SEC) :- time_win(9, 0, 1, @(T, traffic(MES,VAL,SEC)))
 and time_win(8, 0, 1, @(T1, traffic(MES,VAL2,SEC)))
 and MATH(+,RT,T,1)  and COMP(==, RT, T1)
 and COMP(<=, VAL, VAL2) ,

traff_low(MES,SEC) :- time_win(9, 0, 1, diamond(traffic(MES,VAL,SEC)))
 and COMP(>=,VAL, 10)  and COMP(<=,VAL,11),

traff_high(MES,SEC) :- time_win(9, 0, 1, diamond(traffic(MES,VAL,SEC)))
 and COMP(>=,VAL, 250)  and COMP(<=,VAL,300),

poll_low(MES,SEC) :- time_win(9, 0, 1, diamond(pollution(MES,VAL,SEC)))
 and COMP(>=,VAL, 0)  and COMP(<=,VAL,15),

poll_high(MES,SEC) :- time_win(9, 0, 1, diamond(pollution(MES,VAL,SEC)))
 and COMP(>=,VAL, 214)  and COMP(<=,VAL,215),


industrial_area(SEC) :- time_win(9, 0, 1, diamond(poll_inc(MES, SEC)))
 and time_win(9, 0, 1, diamond(traff_dec(MES10,SEC)))
 and time_win(9, 0, 1, diamond(poll_high(MES1,SEC)))
 and time_win(9, 0, 1, diamond(traff_low(MES2,SEC)))
 and COMP(s!=, MES, MES10),

industrial_box(SEC) :- time_win(3, 0, 1, box(industrial_area(SEC))),

highway_area(SEC) :- time_win(9, 0, 1, diamond(traff_inc(MES, SEC)))
 and time_win(9, 0, 1, diamond(poll_inc(MES10,SEC)))
 and time_win(9, 0, 1, diamond(poll_high(MES2,SEC)))
 and time_win(9, 0, 1, diamond(traff_high(MES4,SEC)))
 and COMP(s!=, MES, MES10),

highway_box(SEC) :- time_win(3, 0, 1, box(highway_area(SEC))),

urban_area(SEC) :- time_win(9, 0, 1, diamond(traff_inc(MES, SEC)))
 and time_win(9, 0, 1, diamond(poll_dec(MES10,SEC)))
 and time_win(9, 0, 1, diamond(poll_low(MES2,SEC)))
 and time_win(9, 0, 1, diamond(traff_low(MES4,SEC)))
 and COMP(s!=, MES, MES10),

urban_box(SEC) :- time_win(3, 0, 1, box(urban_area(SEC)))
\end{verbatim}

\subsubsection{LASER code for Query 4}

\begin{verbatim}
traff_inc(MES,SEC) :- time_win(9, 0, 1, @(T, traffic(MES,VAL,SEC)))
 and time_win(8, 0, 1, @(T1, traffic(MES,VAL2,SEC)))
 and MATH(+,RT,T,1)  and COMP(==, RT, T1)
 and COMP(>=, VAL, VAL2) ,

poll_dec(MES,SEC) :- time_win(9, 0, 1, @(T, pollution(MES,VAL,SEC)))
 and time_win(8, 0, 1, @(T1, pollution(MES,VAL2,SEC)))
 and MATH(+,RT,T,1)  and COMP(==, RT, T1)
 and COMP(<=, VAL, VAL2) ,

traff_dec(MES,SEC) :- time_win(9, 0, 1, @(T, traffic(MES,VAL,SEC)))
 and time_win(8, 0, 1, @(T1, traffic(MES,VAL2,SEC)))
 and MATH(+,RT,T,1)  and COMP(==, RT, T1)
 and COMP(<=, VAL, VAL2) ,

traff_low(MES,SEC) :- time_win(9, 0, 1, diamond(traffic(MES,VAL,SEC)))
 and COMP(>=,VAL, 10)  and COMP(<=,VAL,11),

poll_low(MES,SEC) :- time_win(9, 0, 1, diamond(pollution(MES,VAL,SEC)))
 and COMP(>=,VAL, 0)  and COMP(<=,VAL,15),

urban_area(SEC) :- time_win(9, 0, 1, diamond(traff_inc(MES, SEC)))
 and time_win(9, 0, 1, diamond(poll_dec(MES10,SEC)))
 and time_win(9, 0, 1, diamond(poll_low(MES2,SEC)))
 and time_win(9, 0, 1, diamond(traff_low(MES4,SEC)))
 and COMP(s!=, MES, MES10),
\end{verbatim}

\subsubsection{LASER code for Query 5}

\begin{verbatim}
city(3.0):-,
city(4.0):-,
city(6.0):-,
town(8.0):-,
town(10.0):-,
town(1.0):-,
town(2.0):-,
town(5.0):-,
town(7.0):-,
town(9.0):-,

suburb(1.0, 3.0):-,
suburb(2.0, 3.0):-,
suburb(8.0, 4.0):-,
suburb(9.0, 4.0):-,
suburb(10.0, 3.0):-,
suburb(7.0, 6.0):-,
suburb(5.0, 6.0):-,

close(A, C, B) :- suburb(A, B)  and suburb(C, B) and COMP(!=, A, C),

@(T, highPollutionCo(SENS, SEC)) :- time_win(4, 0, 1, @(T, pollution(TYPE, MES, SENS, SEC)))
  and COMP(==, TYPE, 2.0)  and COMP(>, MES, 125),
@(T, highPollutionPM(SENS, SEC)) :- time_win(4, 0, 1, @(T, pollution(TYPE, MES, SENS, SEC)))
  and COMP(==, TYPE, 1.0)  and COMP(>, MES, 125),


highPollutionCo_cont(SENS, SEC) :- time_win(4, 0, 1, box(highPollutionCo(SENS, SEC))),
highPollutionPM_cont(SENS, SEC) :- time_win(4, 0, 1, box(highPollutionPM(SENS, SEC))),

industrialSens(SENS, SEC) :- highPollutionCo_cont(SENS, SEC)  
 and highPollutionPM_cont(SENS, SEC),

industrialSec(SEC) :- industrialSens(SENS1, SEC)  and industrialSens(SENS2, SEC)
  and industrialSens(SENS3, SEC)  and industrialSens(SENS4, SEC)
  and COMP(!=, SENS1, SENS2)  and COMP(!=, SENS1, SENS3)  and COMP(!=, SENS1, SENS4) 
  and COMP(!=, SENS2, SENS3)  and COMP(!=, SENS2, SENS4)  and COMP(!=, SENS3, SENS4),

anomaly(CITY) :- industrialSec(SEC1)
  and industrialSec(SEC2)  and close(SEC1, SEC2, CITY),

industrialArea(SEC) :- industrialSec(SEC)  and city(SEC),

alert(SEC) :- industrialArea(SEC)  and anomaly(SEC)
\end{verbatim}

\subsection{Architectures of the chosen RNNs} 
\paragraph{Query 1:} \\
	LSTM(4 cells, activation=tanh)
	LSTM(4 cells, activation=tanh)
	LSTM(4 cells, activation=tanh)
	LSTM(4 cells, activation=tanh)
	Dense(1 neuron, activation=sigmoid)
	
\paragraph{Query 2:} \\
	LSTM(16 cells, activation=tanh)
	LSTM(10 cells, activation=tanh)
	Dense(10 neuron, activation=sigmoid)
	
\paragraph{Query 3:} \\
	LSTM(8 cells, activation=tanh)
	Dense(4 neuron, activation=softmax)
	
\paragraph{Query 4:} \\
	LSTM(8 cells, activation=tanh)
	LSTM(4 cells, activation=tanh)
	LSTM(2 cells, activation=tanh)
	LSTM(2 cells, activation=tanh)
	Dense(1 neuron)
	
\paragraph{Query 5:} \\
	LSTM(24 cells, activation=tanh)
	LSTM(12 cells, activation=tanh)
	LSTM(6 cells, activation=tanh)
	Dense(3 neuron, activation=sigmoid)

\subsection{Architectures of the chosen CNNs} 

\paragraph{Query 1:} \\
	ConvLayer1D(64 neurons, kernel= 4, activation=relu);  
	ConvLayer1D(64 neurons, kernel= 4, activation=relu);  
	GlobalAveragePooling;
	DenseLayer(64 neurons, activation=relu)
	Dropout(0.4)
	DenseLayer(64 neurons, activation=relu)
	Dropout(0.4)
	DenseLayer(1, activation=sigmoid);

\paragraph{Query 2:} \\
	ConvLayer1D(64 neurons, kernel= 2, activation=relu);  
	ConvLayer1D(64 neurons, kernel= 2, activation=relu);  
	GlobalAveragePooling;
	DenseLayer(32 neurons, activation=relu)
	Dropout(0.6)
	DenseLayer(32 neurons, activation=relu)
	Dropout(0.6)
	DenseLayer(10, activation=sigmoid)
	
\paragraph{Query 3:} \\
	ConvLayer1D(64 neurons, kernel= 2, activation=relu);  
	ConvLayer1D(64 neurons, kernel= 2, activation=relu);  
	GlobalAveragePooling;
	DenseLayer(32 neurons, activation=relu)
	Dropout(0.7)
	DenseLayer(4, activation=softmax)
	
\paragraph{Query 4:} \\
	ConvLayer1D(64 neurons, kernel= 2, activation=elu, padding=valid);  
	ConvLayer1D(64 neurons, kernel= 2, activation=elu, padding=valid);  
	GlobalAveragePooling;
	DenseLayer(64 neurons, activation=elu)
	Dropout(0.6)
	DenseLayer(1 neuron; no activation)
	
\paragraph{Query 5:} \\
	ConvLayer1D(128 neurons, kernel= 2, activation=relu);  
	ConvLayer1D(128 neurons, kernel= 2, activation=relu);  
	GlobalAveragePooling;
	DenseLayer(32 neurons, activation=relu)
	Dropout(0.2)
	DenseLayer(32 neurons, activation=relu)
	Dropout(0.2)
	DenseLayer(3, activation=sigmoid)

\subsection{Training phase results} 
Each graph shows loss and accuracy for training and validation for the indicated number of epochs.

\begin{figure}
\begin{center}
\includegraphics[width=0.8\textwidth]{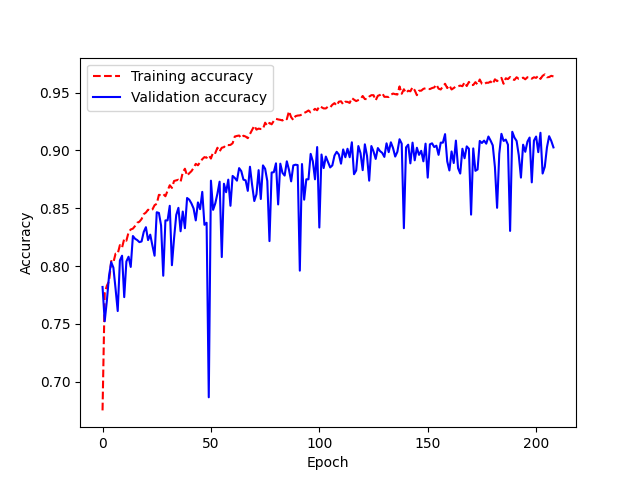}
\caption{Accuracy results of the LSTM training phase for the first Query}
\end{center}
\end{figure}

\begin{figure}
\begin{center}
\includegraphics[width=0.8\textwidth]{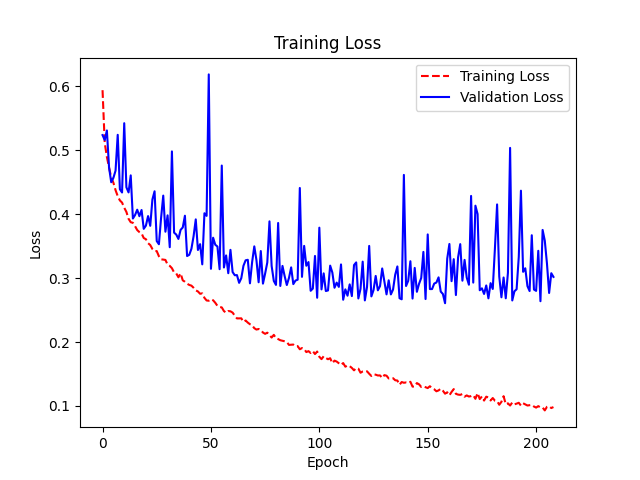}
\caption{Loss results of the LSTM training phase for the first Query}
\end{center}
\end{figure}

\begin{figure}
\begin{center}
\includegraphics[width=0.8\textwidth]{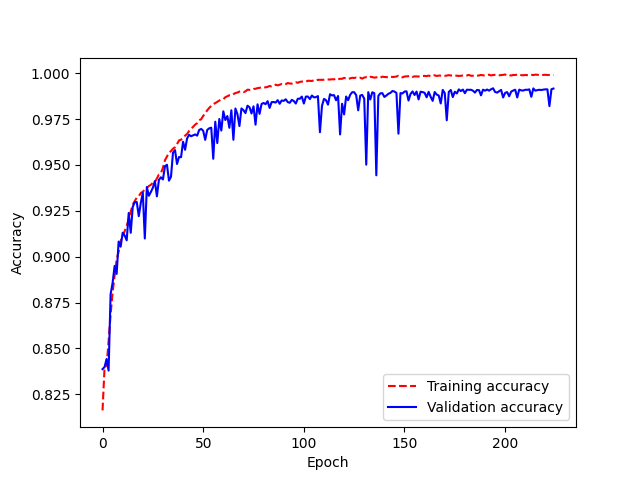}
\caption{Accuracy results of the LSTM training phase for the second Query}
\end{center}
\end{figure}

\begin{figure}
\begin{center}
\includegraphics[width=0.8\textwidth]{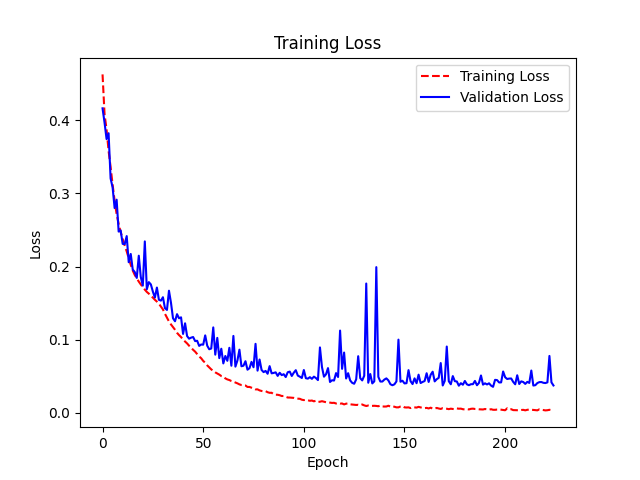}
\caption{Loss results of the LSTM training phase for the second Query}
\end{center}
\end{figure}

\begin{figure}
\begin{center}
\includegraphics[width=0.8\textwidth]{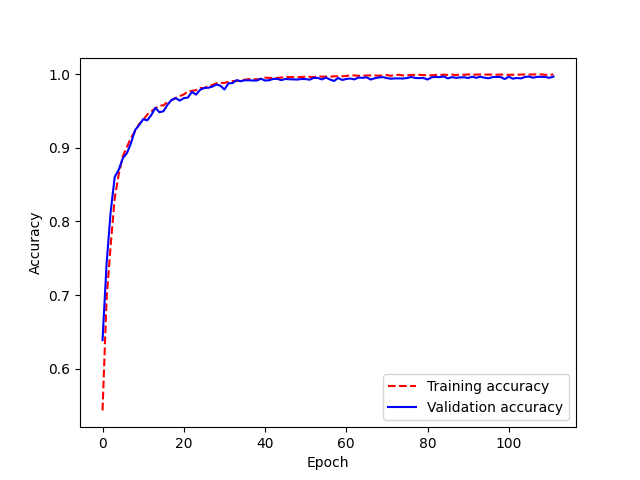}
\caption{Accuracy results of the LSTM training phase for the third Query}
\end{center}
\end{figure}

\begin{figure}
\begin{center}
\includegraphics[width=0.8\textwidth]{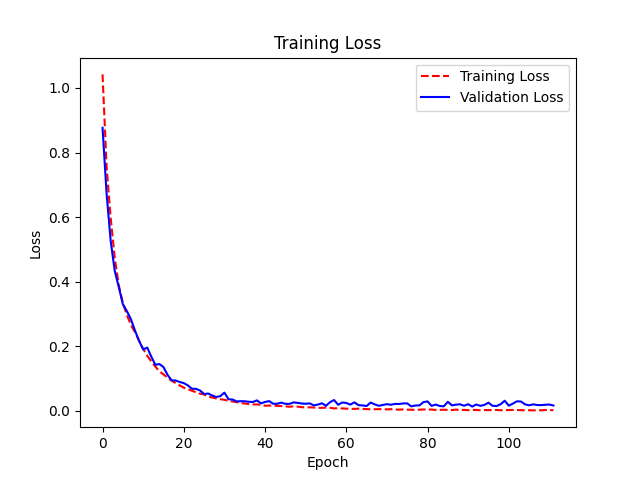}
\caption{Loss results of the LSTM training phase for the third Query}
\end{center}
\end{figure}

\begin{figure}
\begin{center}
\includegraphics[width=0.8\textwidth]{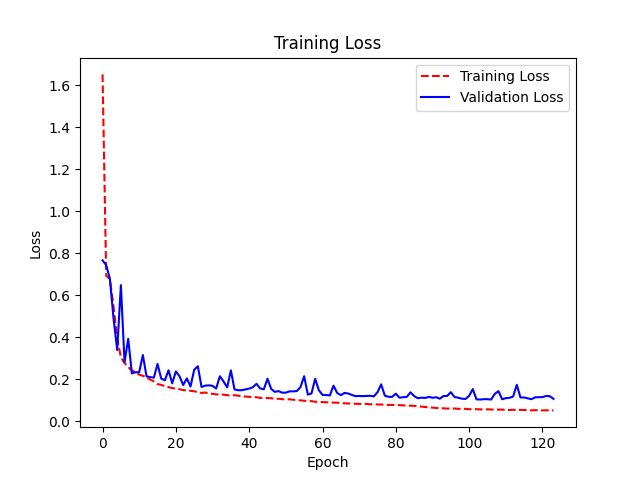}
\caption{MSE results of the LSTM training phase for the fourth Query}
\end{center}
\end{figure}

\begin{figure}
\begin{center}
\includegraphics[width=0.8\textwidth]{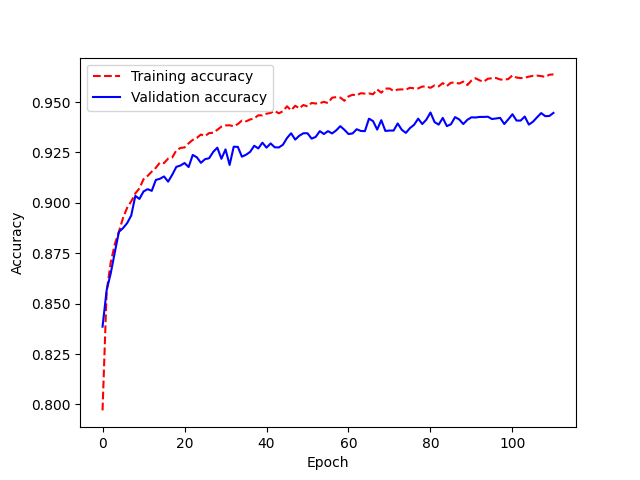}
\caption{Accuracy results of the LSTM training phase for the fifth Query}
\end{center}
\end{figure}

\begin{figure}
\begin{center}
\includegraphics[width=0.8\textwidth]{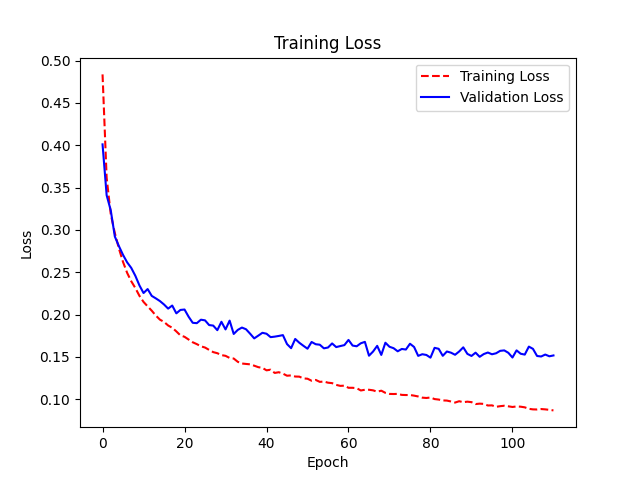}
\caption{Loss results of the LSTM training phase for the fifth Query}
\end{center}
\end{figure}


\begin{figure}
\begin{center}
\includegraphics[width=0.8\textwidth]{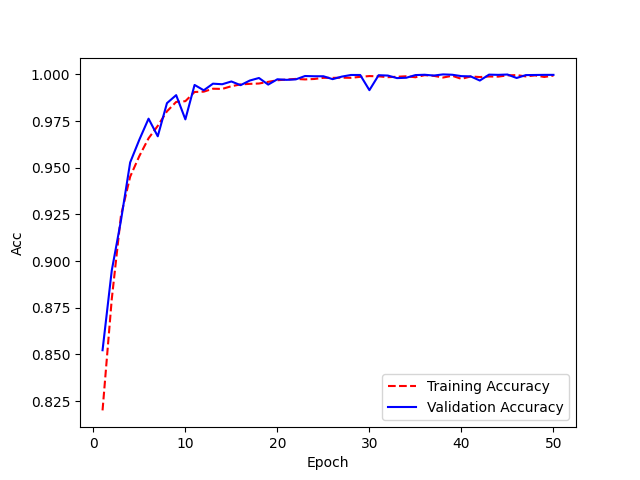}
\caption{Accuracy results of the CNN training phase for the first Query}
\end{center}
\end{figure}

\begin{figure}
\begin{center}
\includegraphics[width=0.8\textwidth]{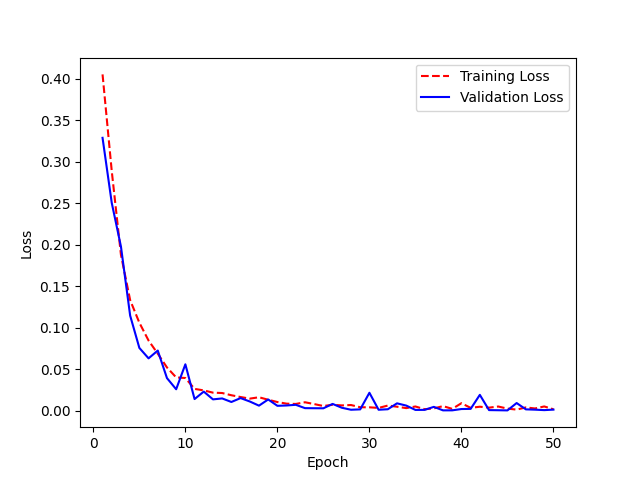}
\caption{Loss results of the CNN training phase for the first Query}
\end{center}
\end{figure}

\begin{figure}
\begin{center}
\includegraphics[width=0.8\textwidth]{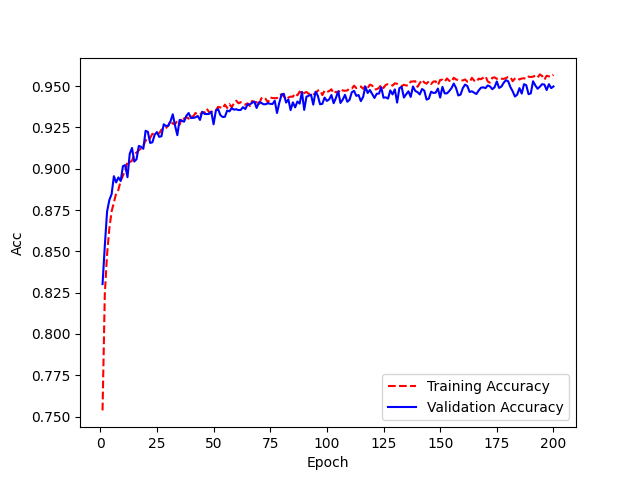}
\caption{Accuracy results of the CNN training phase for the second Query}
\end{center}
\end{figure}

\begin{figure}
\begin{center}
\includegraphics[width=0.8\textwidth]{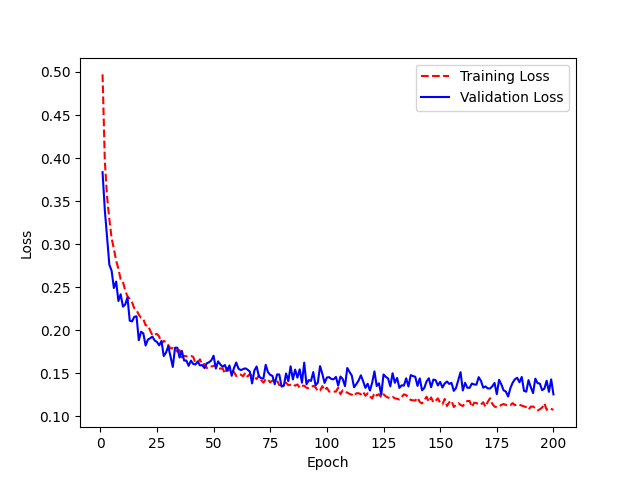}
\caption{Loss results of the CNN training phase for the second Query}
\end{center}
\end{figure}

\begin{figure}
\begin{center}
\includegraphics[width=0.8\textwidth]{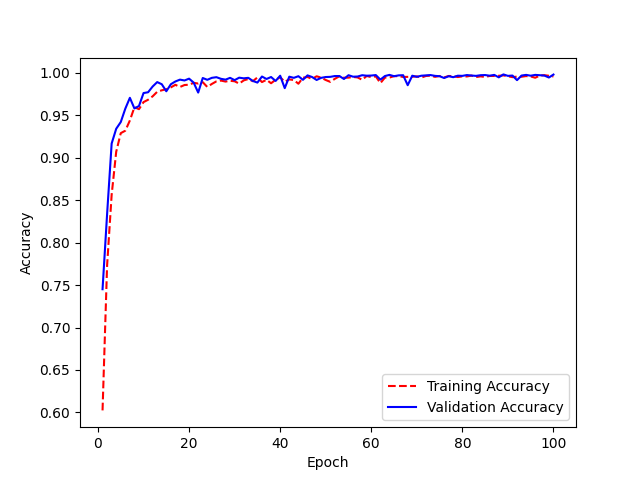}
\caption{Accuracy results of the CNN training phase for the third Query}
\end{center}
\end{figure}

\begin{figure}
\begin{center}
\includegraphics[width=0.8\textwidth]{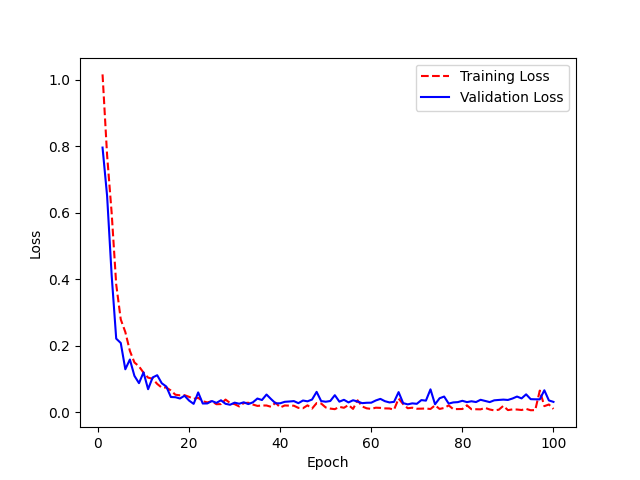}
\caption{Loss results of the CNN training phase for the third Query}
\end{center}
\end{figure}

\begin{figure}
\begin{center}
\includegraphics[width=0.8\textwidth]{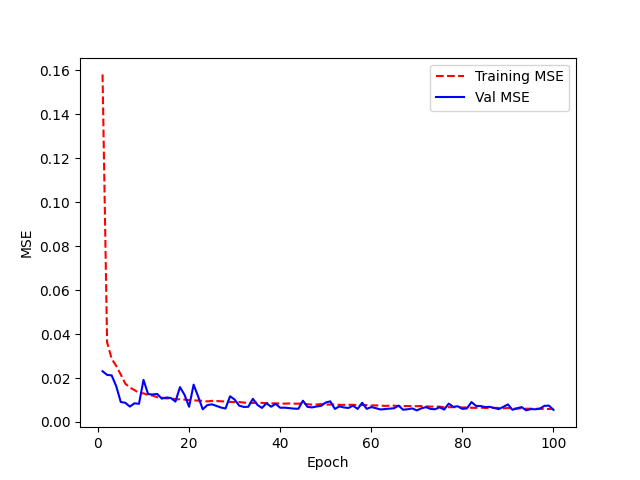}
\caption{MSE results of the CNN training phase for the fourth Query}
\end{center}
\end{figure}

\begin{figure}
\begin{center}
\includegraphics[width=0.8\textwidth]{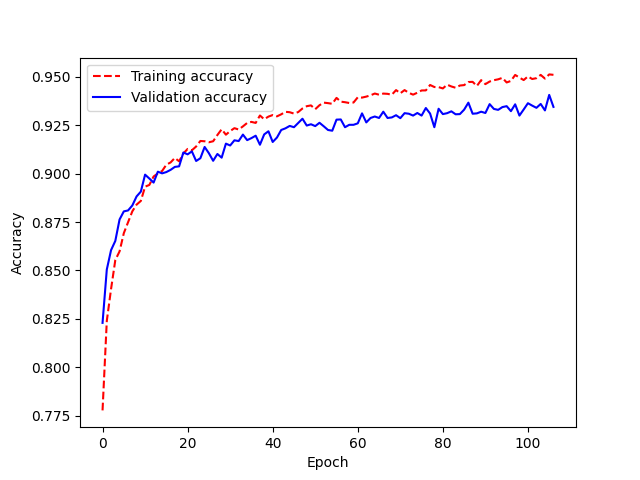}
\caption{Accuracy results of the CNN training phase for the fifth Query}
\end{center}
\end{figure}

\begin{figure}
\begin{center}
\includegraphics[width=0.8\textwidth]{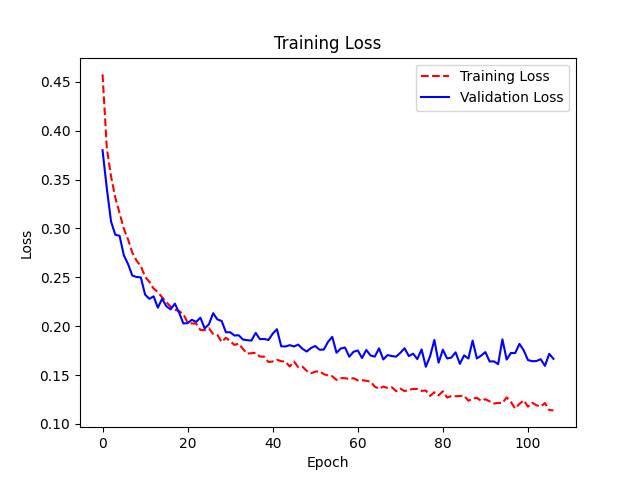}
\caption{Loss results of the CNN training phase for the fifth Query}
\end{center}
\end{figure}

\end{document}